\documentclass[10pt,twocolumn,letterpaper]{article}

\usepackage{cvpr}              % To produce the CAMERA-READY version
\usepackage{graphicx}
\usepackage{amsmath}
\usepackage{amssymb}
\usepackage{booktabs}
\usepackage[utf8]{inputenc}
\usepackage{algorithm}
\usepackage{algpseudocode}

\usepackage[pagebackref,breaklinks,colorlinks]{hyperref}

\usepackage[capitalize]{cleveref}
\crefname{section}{Sec.}{Secs.}
\Crefname{section}{Section}{Sections}
\Crefname{table}{Table}{Tables}
\crefname{table}{Tab.}{Tabs.}

\def\cvprPaperID{*****} % *** Enter the CVPR Paper ID here
\def\confName{CVPR}
\def\confYear{2022}

\begin{document}

\title{Composition Vision-Language Understanding via Segment and Depth Anything Model}

\author{
Mingxiao Huo$^1$, Pengliang Ji$^1$\thanks{Equal Contribution}~, Haotian Lin$^1$, Junchen Liu$^2$, Yixiao Wang$^2$, Yijun Chen$^3$, 
\\ \quad 
$^1$Carnegie Mellon University \quad $^2$ UC Berkeley  $^3$ AWS
}
\maketitle

%%%%%%%%% ABSTRACT
\begin{abstract}
We introduce a pioneering unified library that leverages depth anything, segment anything models to augment neural comprehension in language-vision model zero-shot understanding. This library synergizes the capabilities of the Depth Anything Model (DAM), Segment Anything Model (SAM), and GPT-4V, enhancing multimodal tasks such as vision-question-answering (VQA) and composition reasoning. Through the fusion of segmentation and depth analysis at the symbolic instance level, our library provides nuanced inputs for language models, significantly advancing image interpretation. Validated across a spectrum of in-the-wild real-world images, our findings showcase progress in vision-language models through neural-symbolic integration. This novel approach melds visual and language analysis in an unprecedented manner. Overall, our library opens new directions for future research aimed at decoding the complexities of the real world through advanced multimodal technologies and our code is available at \url{https://github.com/AnthonyHuo/SAM-DAM-for-Compositional-Reasoning}.
\end{abstract}

\section{Introduction}
\label{sec:intro}
\begin{figure*}[t]
  \centering
   \includegraphics[width=1.0\linewidth]{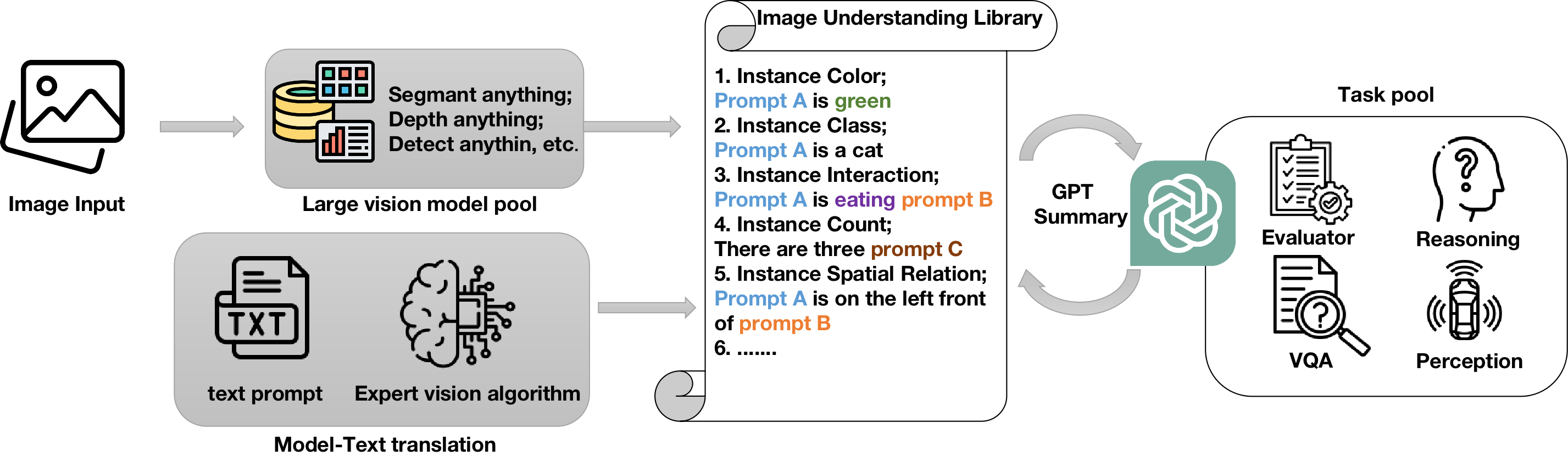}
   \caption{Illustration of the image understanding library. In this paper, we create an library that leverages multi large vision models to extract the rich information for an image input, then using GPT-4V to extract and summary more high level information for the vision language understanding, this pipline can be used in vqa, reasoning and many other tasks.}
  \label{fig:teaser}
\end{figure*}
In the era of large vision models, numerous train-free models have emerged ~\cite{kirillov2023segment,lin2023video}, capable of extracting knowledge from images from various perspectives. Models such as depth anything, detect anything, and segment anything enable the understanding of instance-level intrinsic information, as well as the collection of relative compositional information between different instance-level objects. Leveraging the zero-shot capabilities of these large models allows for the summarization of images at different levels. This process builds a bridge to current large language models, providing summarized text prompts that enable these models to zero-shot understand images. Furthermore, by supplying more accurate image-level information, the results can enhance a variety of multi-modal text-to-image tasks, including vision-question-answering (VQA), reasoning, and scene understanding. This integration signifies a significant advancement in the field, facilitating a deeper understanding of images through language models and improving the efficacy of multi-modal tasks.

In recent works on text-image multi-modal tasks\cite{liu2024visual,liu2023improved,zhu2023minigpt,chen2023minigpt}, the primary focus has been on training specific models to enhance the similarity between text-image pairs and expand multi-modal task capabilities. Presently, numerous large vision models demonstrate the ability to comprehend images. By collecting vision information from images, it is possible to construct a comprehensive image-understanding library, encompassing instance identification, instance classification, instance counting, instance compostional information and more. Subsequently, large language models can retrieve information from this image-understanding library, significantly improving the quality of text-image tasks. This approach not only leverages the strengths of existing vision models but also paves the way for more sophisticated text-image integration, thereby enriching the multi-modal research landscape.

In this paper, we commence by extracting instance-level information through segmentation models, enabling us to capture instance identification along with intrinsic properties such as color. Furthermore, by integrating depth data, we introduce a concept of instance-level depth, which calculates the average depth from current camera viewpoints. Following the acquisition of 2D position information through detection models, we compile comprehensive instance-level compositional information of the entire image. This information is crucial for enhancing the capabilities of current multi-modal tasks. Finally, we validate our image-understanding library's effectiveness by integrating it with previous vision-question-answering tasks. Our results demonstrate its potential to facilitate understanding in complex scenes, thereby underscoring the library's utility in augmenting multi-modal research endeavors.

Additionally, our method incorporates zero-shot compositionality and leverages intrinsic-level understanding, which can significantly benefit autonomous driving and robotics tasks\cite{lin2024joint,huo2023human}. In the context of autonomous driving perception, gathering depth information for different instances can aid in constructing a risk avoidance system. This involves detecting humans and evaluating risk levels through instance depth information. For robotics, our approach can enhance long-horizon tasks in conjunction with large language models (LLMs). Our method's ability to distinguish between different instances based on specific compositions can provide a clearer set of sub-goals for these long-horizon tasks. This clarity in defining sub-goals is crucial for the effective planning and execution of tasks in both autonomous systems and robotic applications, demonstrating the broad applicability and potential impact of our approach.

% \begin{figure}[t]
%   \centering
%   \fbox{\rule{0pt}{2in} \rule{0.9\linewidth}{0pt}}
%    %\includegraphics[width=0.8\linewidth]{egfigure.eps}

%    \caption{Example of caption.}
%    \label{fig:onecol}
% \end{figure}

\section{Related Works}
Leveraging a unified model effectively in the large language model era is crucial for various tasks, particularly for zero-shot tasks. Inspired by prior successful works such as Depth Anything\cite{yang2024depth} and Segment Anything\cite{kirillov2023segment} to address complex compositional problems in the wild tasks by combining the strong open-set foundation models, two exemplary unified models have emerged. These models exemplify the strategic application of unified approaches to tackle complex problems in depth perception and image segmentation, demonstrating the potential for broad applicability and efficiency in processing diverse tasks without the need for task-specific training.
\label{sec:formatting}
\subsection{Unified Model}
Depth Anything Model (DAM)\cite{yang2024depth} presents a novel approach to monocular depth estimation leveraging large-scale unlabeled data. It introduces a data engine for automatic annotation of a vast dataset, enhancing model generalization across diverse scenes. The methodology includes challenging optimization targets and auxiliary supervision from pre-trained encoders, significantly improving zero-shot depth estimation capabilities. The paper sets new standards in depth estimation, demonstrating substantial advancements over existing models through extensive evaluation on multiple datasets.

DAM enhances zero-shot task performance in depth estimation, enabling it to predict depth maps for scenes or objects not seen during training. Additionally, it supports compositional knowledge learning, allowing the model to generalize depth estimation across various contexts by integrating learned depth cues and features, showcasing an advanced understanding of spatial relationships and object geometries beyond specific training instances.

Segment Anything Model (SAM)\cite{kirillov2023segment} introduces a comprehensive approach to image segmentation, creating a model and dataset that enable zero-shot transfer to new tasks and distributions. They developed a large dataset with over 1 billion masks from 11 million images and a model capable of real-time, prompt-based segmentation. This work represents a significant leap in segmentation research, offering a foundation model for vision that adapts flexibly to various segmentation tasks through prompt engineering. SAM embodies a significant stride in zero-shot task capabilities by facilitating accurate segmentation of previously unseen objects, leveraging extensive visual concept learning. Furthermore, it advances compositional knowledge learning, enabling it to interpret and segment based on novel, intricate prompts through the synthesis of learned concepts. This denotes a profound understanding of visual data, surpassing mere exposure during training. 

With SAM and DAM for compositional knowledge learning signifies a pioneering approach in computer vision, where the integrated model leverages the strengths of both segmentation and depth estimation. This amalgamation enables a richer understanding of visual scenes, enhancing the model's ability to interpret and analyze images with nuanced depth and segmentation details, facilitating advanced compositional knowledge across diverse visual tasks.

\section{Methods}

Our image-understanding library comprises three primary components. Firstly, we distinguish instance-level intrinsic properties of instances with bounding boxes and masks instructed by user-defined text prompts. Secondly, we extract and determine the depth of user-interested instances. Lastly, we integrate knowledge to analyze the extrinsic information represented by compositional relationships via composition reasoning modules.

\subsection{Symbolic Knowledge Gathering}
\label{3-1}
The goal of our pipeline is to extract instance-level intrinsic and extrinsic properties i.e. symbolic knowledge from in-the-wild images. A typical approach brings 3D geometry to achieve this. However, directly predicting the geometry of instances in a 2D in-the-wild image is challenging and imprecise. Inspired by the assumption that sensors widely adopted by autonomous driving can mimic spatial information through detection and depth cues, and that segmentation can indicate potential instance-level interaction, we aim to gather instance-level information to facilitate image understanding by integrating these multimodal signals. To this end, we propose a simple, effective, and computationally feasible generic zero-shot framework to acquire compositional knowledge using unified models.

\begin{figure*}[t]
  \centering
   \includegraphics[width=\linewidth]{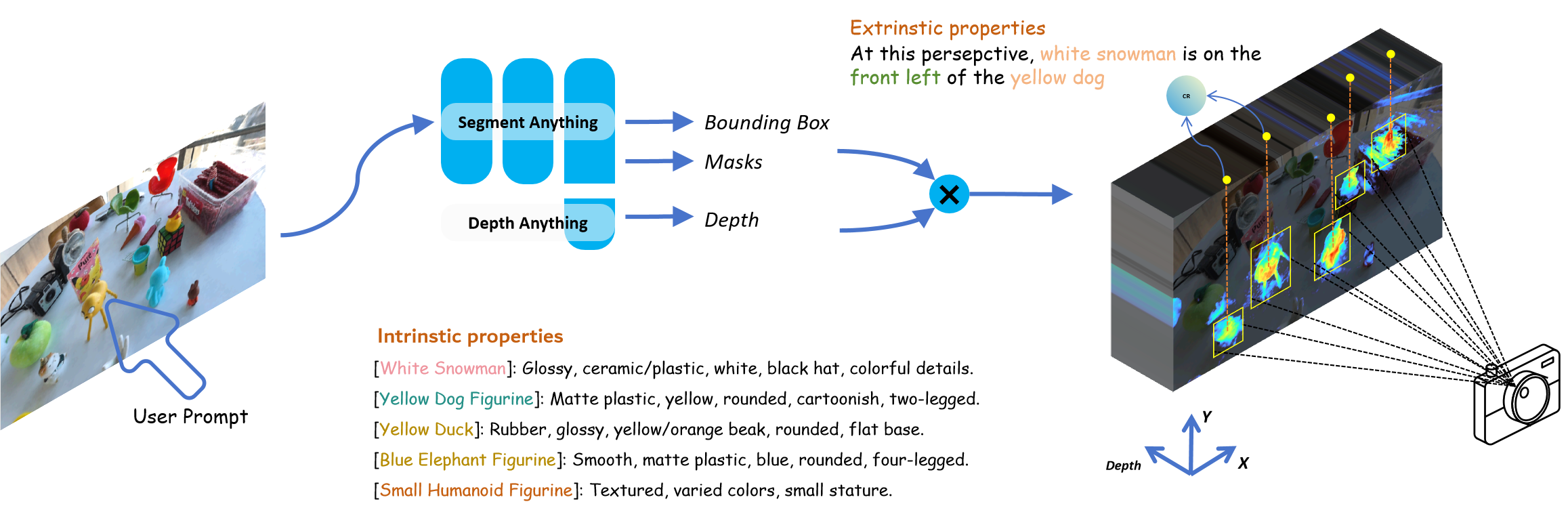}
   \caption{Illustration of the synergy between the Depth Anything Model (DAM) and Segment Anything Model (SAM) in enhancing understanding of images captured in natural settings. By utilizing SAM, we differentiate between instances through the application of masks and bounding boxes, while DAM is employed to produce 2.5D depth information, facilitating the comprehension of 3D signals and contributing to a richer understanding of scene composition. Ultimately, by integrating the insights from GPT-4V, we construct a comprehensive response to queries or delineate the intrinsic and extrinsic characteristics of objects of interest using text prompts.}
  \label{fig:short}
\end{figure*}
In our library, given an input in-wild 2D image, we first leverage SAM\cite{kirillov2023segment} to generate precise mask annotations for each instance within the image. Subsequently, for each instance, we synchronize the pixel-wise depth results obtained from DAM\cite{yang2024depth} with the corresponding instance, utilizing an averaging function. This process aggregates depth values associated with the instance to produce a consolidated depth measurement for each instance. As illustrated in Fig. 2, our model provides compositional knowledge with these signals for each pair of instances based on text prompt input in both conventional and long-tail scenarios.

To further enhance scalability and adaptability, our framework is designed to accommodate additional outputs from various large-scale vision models. This integration facilitates the enrichment of instance-level knowledge with advanced insights, such as action prediction and human pose estimation. Such enhancements significantly elevate the understanding of symbolism within the framework, paving the way for more nuanced and sophisticated image interpretation capabilities.

\subsection{Composition Reasoning}
The general definition of compositional knowledge encompasses relative spatial information, such as "at the top of," and interactive information, such as "inside," from the perspective of the input camera. In this paper, we concentrate on $8$ fundamental spatial relations and $6$ complex interaction types. Utilizing symbolic knowledge gathered, as indicated in section \ref{3-1}, the composition reasoning module generates output by considering the knowledge of location, distance, and size between pairs of instances. The detailed algorithm of our composition reasoning module is shown in Table \ref{tab1}.

\begin{table}
  \centering
  \begin{tabular}{@{}cc@{}}
    \toprule
    Dimension & Compositional Knowledge \\
    \midrule
    Spatial-X & Left and Right \\
    Spatial-Y & At the top of, at the bottom of \\
    Spatial-Z & Front and Back \\
    Advanced & Inside, Beside,  \\
    \bottomrule
  \end{tabular}
  \caption{Compositional Text Setting.}
  \label{tab1}
\end{table}

Regarding compositional knowledge, we prioritize spatial information as the basis for representation and introduce a discriminative composition reasoning algorithm at the level of instance pairs, as delineated in Algorithm 1. By exploiting the capabilities of these advanced generative models, we are able to extract detailed information from merely a two-dimensional image captured in natural settings.

\begin{algorithm}
\caption{Composition Reasoning Module}
\begin{algorithmic}[1]
\State Input: in-wild images $I$, optional text prompt $\psi$.
\State Output: composition text of instances $O$.
\State $(mask_i, class_i)_{i \in objs} \gets SAM(I,\ prompt=\psi)$
\State $(depth_j)_{j \in pixs} \gets DAM(I)$
\State $(depth_i)_{i \in objs} \gets AVG(depth_j)_{j \in mask_i, i \in objs}$
\For{$y_i \in \mathcal{Y}$ and $(txt_i, img_i) \in D$}
\EndFor
\end{algorithmic}
\label{alg1}
\end{algorithm}

\subsection{Visual Language Answering Boosting}
We present an exemplary application facilitated by our library, demonstrating its utility in the domain of zero-shot Visual Question Answering (VQA) with Large Language Models (LLMs), particularly using GPT-4. Our library enhances GPT-4's ability to generate accurate answers to complex, real-world questions about images. For instance, questions such as "How many white dogs are playing with each other in this image?" or "What is the relationship between the humanoid toy and the white box?" are addressed by generating responses from GPT-4, which are informed by a well-structured prompt. This prompt integrates the question with instance-level image knowledge, denoted as $C$, derived from previous analyses.

Initially, we enhance the prompt's design for GPT-4, incorporating both intrinsic knowledge—such as color and quantity—and extrinsic knowledge, which includes relational information between objects of interest. This enriched instance-level knowledge comprises both neural (semantic) and symbolic (spatial) information, creating a comprehensive and instructive prompt. For example, a revised prompt might be, "Considering the original question about the number of toys in the boxes, we have ascertained that the smallest toy is positioned behind the large green toy on the right (auxiliary knowledge)." Such detailed input enables GPT-4 to process the information more accurately. In contrast, without these symbolic cues, a large multimodal model might struggle to integrate symbolism with neural information, potentially overlooking crucial details like a small blue toy that is difficult to identify at the feature level.

By synergizing visual information from a large vision model with linguistic insights from a large language model—and merging these at the symbolic level—our library achieves precise outcomes for a broader range of scenarios. This is made possible by our two-branch approach, which enhances generalizability and accuracy.

Furthermore, our library aids in narrowing the focus of large vision models to objects of interest within a question by utilizing textual information as a condition for Spatially Aware Modeling (SAM). Consequently, it assesses the relationships between selected object pairs, thereby enhancing the library's scalability and practicality for diverse applications.

\section{Experiments}

\subsection{Zero-shot Composition Reasoning}
\begin{figure}[t]
  \centering
   \includegraphics[width=\linewidth]{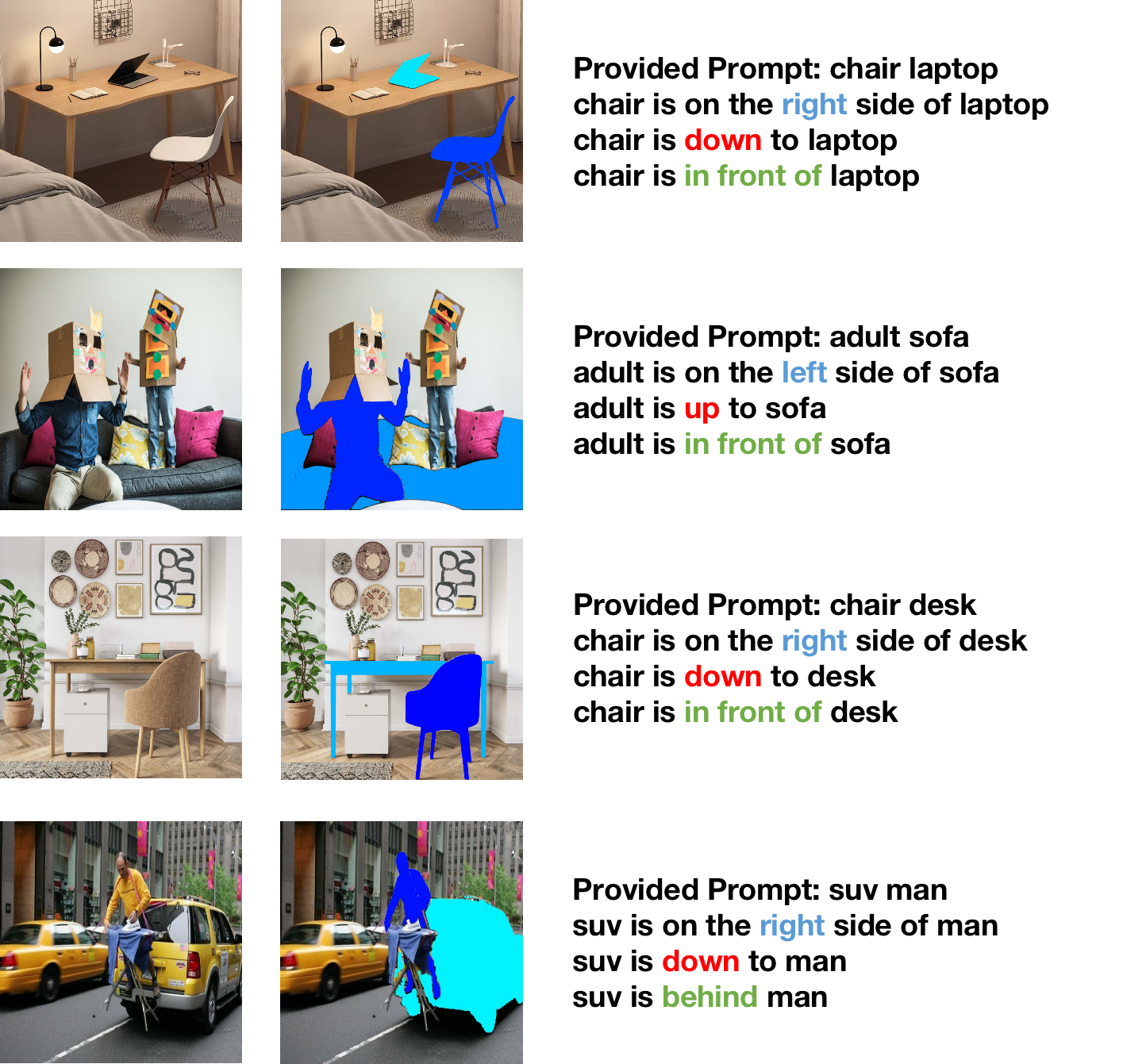}
   \caption{Quantitative results on in-the-wild images for composition reasoning. }
  \label{fig:3}
\end{figure}

We demonstrate the ability of our library for images in different scenarios. The detailed quantitative results are shown in Figure \ref{fig:3}.

In this experiments, for the wild images we provide some prompts for the object of interest in one scene, and getting forward the SAM and DAM and detect anything model, the image understanding library shows the ability to output the spatial relationship between the objects in the scene. As shown in Fig.~\ref{fig:3}, leverage the semantic information from the sam model, and the 2d perception information from detect anything model, also the 3d perception ability from the dam model, our library can show the zero-shot ability to gather the spatial relationship between two prompted-objects.

\subsection{Zero-shot Symbolic Visual Question Answering}

\begin{figure}[t]
  \centering
   \includegraphics[width=\linewidth]{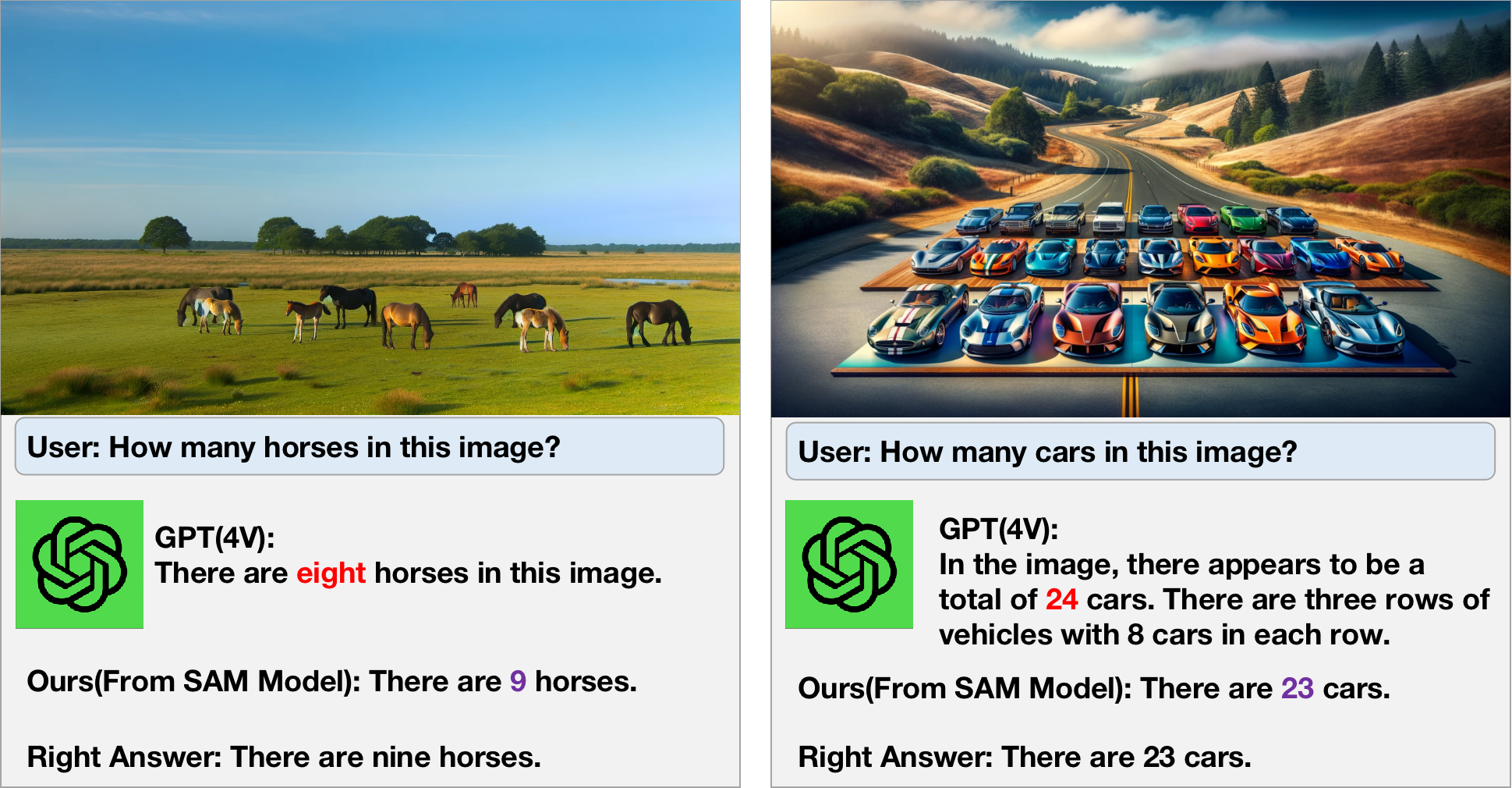}
   \caption{The ability of zero-shot symbolic visual question answering for image understanding library, leveraging the ablities of large vision models, image understanding can outperforms GPT-4V on understanding the symbolic information of one scene, like counting numbers of one object.}
  \label{fig:4}
\end{figure}

In this section, we do experiments about testing the ability for the zero-shot symbolic visual question answering as shown in Fig. ~\ref{fig:4}.

Zero-shot symbolic visual question answering (VQA) leverages the advanced capabilities of large vision models to enhance image understanding, particularly in interpreting symbolic information within a scene. Unlike traditional models such as GPT-4V, zero-shot symbolic VQA excels in tasks like counting objects and recognizing symbolic elements without requiring prior training on specific datasets. By utilizing pre-trained models that generalize across various tasks, this approach can accurately answer questions about new images, demonstrating superior performance in tasks involving detailed symbolic interpretation and object counting.

\subsection{Enhancing Vision-language Understanding}

The enhancement of vision-language models, such as GPT-4V, can be significantly achieved by integrating an image understanding library that specializes in extracting spatial information from images. This library can dissect an image into its spatial components, identifying relationships and locations of objects within the scene. When this detailed spatial information is combined with the language processing capabilities of GPT-4V, the model can achieve a more nuanced and comprehensive understanding of the image. This integration allows the vision-language model to interpret and express complex spatial relationships more accurately, thereby enhancing its performance in tasks that require a sophisticated understanding of spatial dynamics and interactions within a scene.

\section{Discussion and Conclusion}
In this paper, we embarked on addressing the prevailing limitations inherent in current vision-language understanding frameworks by introducing a unified library that leverages integrated models to access and align symbolic information. This novel approach was rigorously evaluated through its application to two multimodal tasks: visual-question answering and composition reasoning. Our qualitative assessment, based on in-the-wild images, has highlighted the potential of our method.

Our innovative framework paves the way for the generation of a substantial in-the-wild image-compositional knowledge dataset, serving as a rich training resource. We extend an invitation to the research community to employ our pipeline for dataset creation, thereby enhancing their generative models, such as visual-language models (VLMs), with a nuanced understanding of compositional knowledge.

Despite the proficiency of current visual-language models in interpreting objects within an image for tasks such as Visual Question Answering (VQA), the intricate compositional knowledge between two arbitrary objects remains a challenge—often being inefficient, unstable, and costly to acquire through complex prompts. Our methodology introduces a streamlined approach to augment visual question answering capabilities without necessitating extensive training, merely through refinement based on simple prompts within the visual embedding space.

Nonetheless, it is imperative to acknowledge that the efficacy of our library is significantly dependent on the advancements within multimodal large language models (LLMs). We anticipate that as LLMs evolve, our library will correspondingly become more stable, general, and reliable.

Looking ahead, we remain committed to integrating the latest developments in open-source unified models into our library. This ongoing effort forms a cornerstone of our future work, ensuring that our library remains at the forefront of facilitating advanced vision-language understanding. This endeavor not only addresses current limitations but also opens new avenues for research and development in the quest for a deeper comprehension of real-world complexities through the lens of advanced multimodal technologies.

\begin{figure*}[t]
  \centering
   \includegraphics[width=1.0\linewidth]{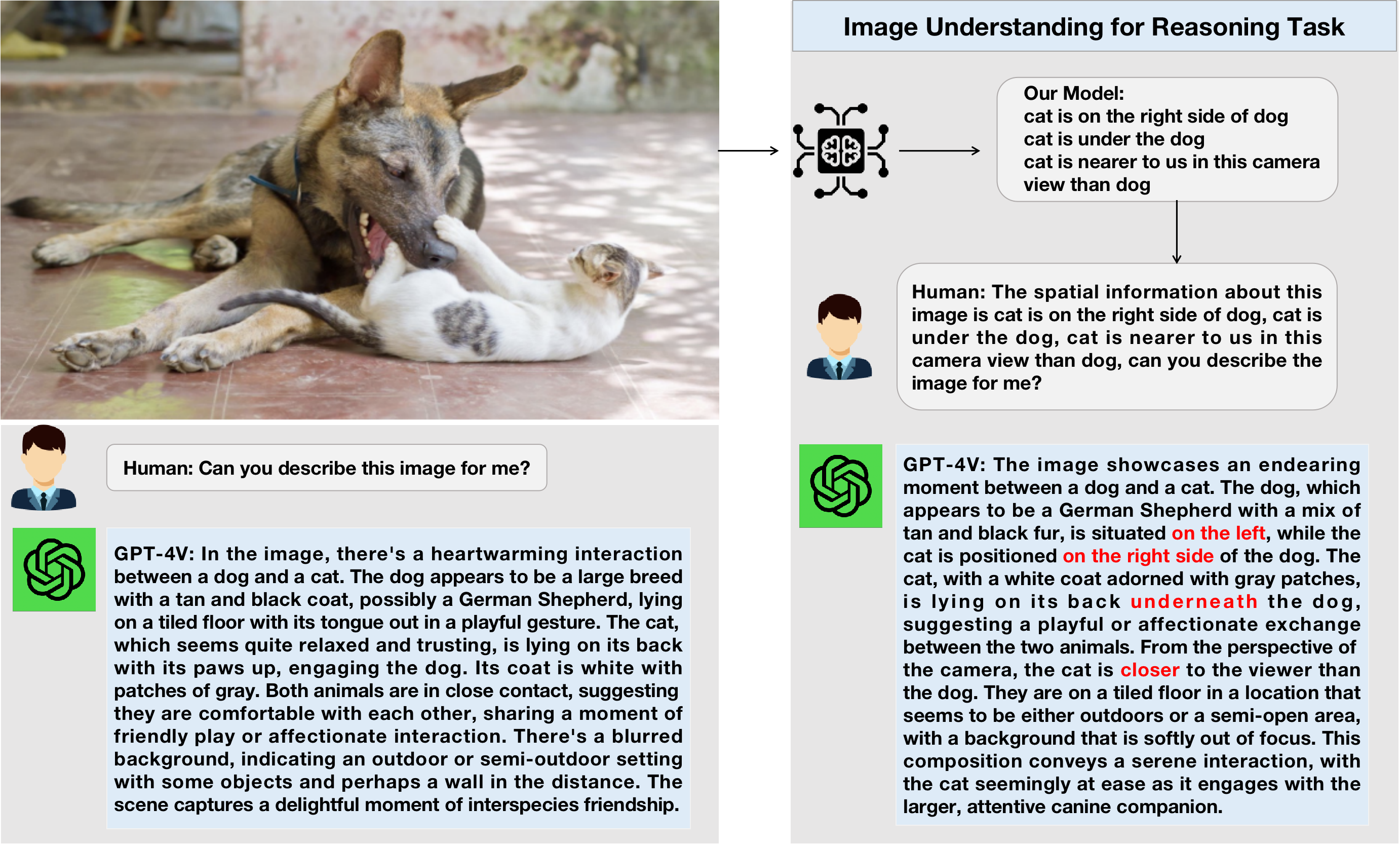}
   \caption{Illustration of the enhancement for the vision-language model by using the image understanding library. The image understanding library can extract the spatial information from an image, then after combining the gathered spatial information, the GPT-4V can enhance the original vision-language understanding with more complex spatial information expression.}
  \label{fig:gpt}
\end{figure*}

% In this paper, to address the shortcomings of existing vision-language understanding, we introduce a library integrated with unified models, a novel approach that access symbolic information by considering and deal their alignment. To evaluate our method, we employ our model on two multimodal task including visual-question answering and composition reasoning. The qualitive results on in-the-wild images shows . 

% Our framework facilitates the creation of a large-scale in-the-wild image-compositional knowledge dataset as training data. We welcome future research utilizing our pipeline to generate datasets, equipping their generative models, such as visual-language models (VLMs), with a general understanding of composition.

% Though existing visual-language models in the Visual Question Answering (VQA) task can understand objects well in an image, the compositional knowledge of two arbitrary objects in the visual-language embedding space could be inefficient, unstable, and expensive to get from complex prompts. With our method, we could efficiently improve visual question answering without training, by just refining on the answer with a simple prompt in the visual embedding space. 

% However, the current capabilities of our library highly replies on multimodal large language models. We anticipate that as LLMs become more
% advanced, our library will also become increasingly
% stable, genearl and reliable

% As new open-source unified models continue to emerge, we will keep track of the latest developments and incorporate their into our library as part of our future efforts.

{\small
\bibliographystyle{ieee_fullname}
\bibliography{egbib}
}

\end{document}